\documentclass{article}

\usepackage{microtype}
\usepackage{graphicx}
\usepackage{subcaption}
\usepackage{booktabs} %

\usepackage{hyperref}

\usepackage[preprint]{icml2026}

\usepackage{amsmath}
\usepackage{amssymb}
\usepackage{mathtools}
\usepackage{amsthm}
\usepackage{bm}

\usepackage[normalem]{ulem}

\usepackage[capitalize,noabbrev]{cleveref}
\setcitestyle{numbers,square}
\theoremstyle{plain}

\theoremstyle{definition}

\theoremstyle{remark}

\usepackage[textsize=tiny]{todonotes}

\usepackage[dvipsnames]{xcolor}

\usepackage{xcolor}         %
\usepackage{colortbl}

\icmltitlerunning{OV-DEIM: Real-time DETR-Style Open-Vocabulary Object Detection with GridSynthetic Augmentation}

\begin{document}

\twocolumn[
  \icmltitle{OV-DEIM: Real-time DETR-Style Open-Vocabulary Object Detection with GridSynthetic Augmentation}

  \icmlsetsymbol{equal}{*}

  \begin{icmlauthorlist}
    \icmlauthor{Leilei Wang}{comp,yyy,xxx}
    \icmlauthor{Longfei Liu}{comp}
    \icmlauthor{Xi Shen}{comp}
    \icmlauthor{Xuanlong Yu}{comp}
    \icmlauthor{Ying Tiffany He}{xxx}
    \icmlauthor{Fei Richard Yu}{xxx,sch}
    \icmlauthor{Yingyi Chen}{zzz}
  \end{icmlauthorlist}

  \icmlaffiliation{comp}{Intellindust AI Lab} 
  
  \icmlaffiliation{yyy}{Guangdong Laboratory of Artificial Intelligence and Digital Economy (SZ), Shenzhen, China} 
  
  \icmlaffiliation{xxx}{College of Computer Science and Software Engineering, Shenzhen University, China} 
  
  \icmlaffiliation{sch}{School of Information Technology, Carleton University, Canada} 
  \icmlaffiliation{zzz}{Institute for Research in Biomedicine, Bellinzona, Switzerland}
  \icmlcorrespondingauthor{Fei Richard Yu}{yufei@szu.edu.cn}
  \icmlcorrespondingauthor{Yingyi Chen}{yingyi.chen@irb.usi.ch}

  \icmlkeywords{Real-time Open-Vocabulary Object Detection, DETR}

  \vskip 0.3in
]

\printAffiliationsAndNotice{}  %

\begin{abstract}
  Real-time open-vocabulary object detection (OVOD) is essential for practical deployment in dynamic environments, where models must recognize a large and evolving set of categories under strict latency constraints. Current real-time OVOD methods are predominantly built upon YOLO-style models. In contrast, real-time DETR-based methods still lag behind in terms of inference latency, model lightweightness, and overall performance. In this work, we present OV-DEIM, an end-to-end DETR-style open-vocabulary detector built upon the recent DEIMv2 framework with integrated vision–language modeling for efficient open-vocabulary inference. We further introduce a simple query supplement strategy that improves Fixed AP without compromising inference speed. Beyond architectural improvements, we introduce GridSynthetic, a simple yet effective data augmentation strategy that composes multiple training samples into structured image grids. By exposing the model to richer object co-occurrence patterns and spatial layouts within a single forward pass, GridSynthetic mitigates the negative impact of noisy localization signals on the classification loss and improves semantic discrimination, particularly for rare categories. Extensive experiments demonstrate that OV-DEIM achieves state-of-the-art performance on open-vocabulary detection benchmarks, delivering superior efficiency and notable improvements on challenging rare categories. Code and pretrained models are available at : \url{https://github.com/wleilei/OV-DEIM}. 
\end{abstract}

\section{Introduction}

Real-time object detection is an important task in computer vision, underpinning numerous applications such as image understanding, robotics, and autonomous driving. 
Recent advances in both YOLO-style models~\cite{redmon2016you,yolox,yolov4} and DETR-style models~\cite{detr,deim,rt-detr} have led to significant progress in detection performance. However, these methods are inherently limited to fixed-vocabulary settings, such as the 80 predefined categories in the COCO~\cite{coco} dataset. Once object categories are predefined and annotated, trained detectors can only recognize those categories, which limits their generalization capability and applicability in open-world scenarios~\cite{yolo-world}.

To address this, Open-Vocabulary Object Detection (OVOD) has been proposed, which leverages a shared vision-text space learned from large-scale image-text data to enable recognition of unseen object categories~\cite{ov-dino,yoloe,yolo-world,grounding-dino,glip}. 
However, this advantage comes with two major challenges, namely maintaining high inference efficiency for real-time applications and achieving robust recognition performance on rare or long-tail categories.

YOLO-World~\cite{yolo-world} adopts the standard YOLO architecture~\cite{yolov8} and encodes input texts into an offline vocabulary using a pretrained CLIP~\cite{clip} text encoder. Rather than relying on large and computationally heavy detectors~\cite{ov-dino,glip}, it employs a lightweight detector with a cross-modality fusion network to fuse text and image features, enabling efficient inference over the offline vocabulary without re-encoding textual prompts. Furthermore, YOLOE~\cite{yoloe} demonstrates that the cross-modality fusion can be removed to avoid costly visual-textual feature interaction with diverse negative prompts~\cite{yao2024detclipv3}. Nevertheless, both YOLO-World and YOLOE rely on dense one-to-many assignment and require Non-Maximum Suppression (NMS) during post-processing to eliminate duplicate predictions. This additional heuristic step introduces extra inference latency~\cite{rt-detr}. More importantly, despite achieving competitive overall performance, their recognition accuracy on rare categories remains substantially lower than that on frequent categories in open-vocabulary settings, highlighting limitations in handling long-tail distributions.

To address these issues, we propose OV-DEIM, a strong real-time open-vocabulary object detector. Built upon the real-time DETR-based framework DEIMv2~\cite{deimv2}, OV-DEIM preserves its end-to-end set prediction design and efficient architecture, thereby eliminating the need for NMS while maintaining low-latency inference. Specifically, by adopting different backbone configurations across model scales, OV-DEIM employs DINOv3~\cite{dinov3} for the larger variant and DINOv3-distilled Tiny ViTs~\cite{dosovitskiy2020image} for smaller ones. In this way, OV-DEIM leverages large-scale pretrained visual knowledge while preserving computational efficiency. Combined with the structured inductive bias of transformer-based set prediction, this design strengthens semantic discrimination for rare and long-tail categories, achieving a favorable balance between efficiency and recognition robustness in open-vocabulary settings.

To further improve scalability, we introduce a lightweight query supplement strategy that avoids introducing a large number of additional decoder queries, which would otherwise increase computational cost. This design improves Fixed AP~\cite{FixedAP} without sacrificing inference speed. 

Motivated by the need to increase positive matches in DEIM~\cite{deim}, we further introduce GridSynthetic, a grid-based data augmentation strategy that generates synthetic training images through structured composition. We first construct an object pool by extracting annotated objects from the original dataset, with slightly expanded bounding boxes to preserve contextual information. 
For each synthetic sample, multiple object patches are arranged into an $m \times n$ grid, producing a composite image with updated bounding boxes and labels. Optionally, two synthetic images are blended to further enhance diversity. This structured composition enriches object co-occurrence patterns and improves classification robustness. Moreover, by simplifying spatial layouts, GridSynthetic reduces localization difficulty and encourages diverse cross-category combinations, facilitating stronger semantic relationship learning. Compared with Copy-Paste~\cite{copy_paste}, it provides more effective instance-level supervision, and it can be seamlessly combined with MixUp~\cite{mixup_od} for additional performance gains.

Experimentally, OV-DEIM demonstrates promising performance. Pretrained on Object365V1~\cite{o365}, GQA~\cite{gqa}, and Flickr30k~\cite{flickr30k}, our approach achieves competitive zero-shot results on both LVIS~\cite{lvis} and COCO~\cite{coco}, while maintaining significantly lower inference latency. In particular, on the challenging rare categories of LVIS~\cite{lvis}, our S-, M-, and L-sized models outperform the corresponding YOLOE models (using YOLOv8~\cite{yolov8}) by 4.6 AP, 1.7 AP, and 3.5 AP, respectively. These results demonstrate that OV-DEIM achieves a better balance between efficiency and performance, especially for difficult and long-tail categories.

To summarize, our main contributions are as follows:

\begin{itemize}
\item We propose OV-DEIM, a real-time DETR-style framework for open-vocabulary object detection. By leveraging direct set prediction and a lightweight query design, our method removes category-dependent post-processing and achieves strong zero-shot performance while maintaining high inference efficiency. It achieves a promising balance between speed and accuracy compared to existing approaches.

\item We introduce GridSynthetic, a simple yet effective grid-based data augmentation strategy that improves classification supervision by increasing object diversity and cross-category combinations. It reduces the impact of noisy localization signals and strengthens semantic robustness, especially for rare categories, without adding any inference cost.
\end{itemize}

Our code and pretrained models are available at \url{https://github.com/wleilei/OV-DEIM}.

\section{Related Work}

\paragraph{Open-Vocabulary Object Detection (OVOD)} extends detection beyond fixed taxonomies via large-scale vision-language pre-training. Early OVOD systems transfer image-text alignment into detectors through grounding-based and CLIP-style training, including the GLIP series~\citep{glip, glipv2}, DetCLIP series~\cite{ yao2024detclipv3,yao2022detclip,yao2023detclipv2}, and OWL-ViT series~\citep{minderer2022simple, minderer2023scaling}. In parallel, DETR-style OVOD methods inject language-guided queries into transformer detectors, such as OV-DETR~\cite{zang2022open}, OV-DINO~\cite{wang2024ov}, and GroundingDINO~\cite{grounding-dino}, with the Grounding DINO family further scaling training on large grounding corpora~\citep{GDINO1-5}.

Recent efforts focus on lightweight and real-time OVOD. YOLO-World~\cite{yolo-world} and YOLOE~\cite{yoloe} incorporate vision–language alignment into efficient YOLO architectures, providing a favorable trade-off between accuracy, efficiency, and deployability in practice. 
However, their reliance on NMS can hinder runtime efficiency. DETR-based real-time OVOD, \textit{e.g.}, Grounding DINO 1.5 Edge~\cite{GDINO1-5}, removes NMS and improves latency, yet typically lags behind strong YOLO-based models in accuracy.
To address this gap, we propose OV-DEIM, advancing real-time DETR-style OVOD to achieve higher detection accuracy with lower latency than YOLO-style OVOD, improving practicality for deployment.

\paragraph{Data Augmentations in Object Detection} Data augmentation are widely used to diversify the training sample space~\cite{crowded_copy_paste}. Beyond standard augmentation techniques, \textit{e.g.}, color jittering, random affine transformations, and random flipping, recent detectors such as YOLOv8~\cite{yolov8} and YOLOE~\cite{yoloe} adopt mosaic augmentation~\cite{yolov4}, which composes each input from four rescaled image patches and effectively increases instance density. In DETR-style detectors, DEIM~\cite{deim} demonstrates that model performance can be improved by increasing the number of targets for dense supervision, and accordingly introduces Copy-Paste~\cite{copy_paste} and MixUp~\cite{mixup_od} augmentations. However, Copy-Paste augmentation is inherently constrained in scaling up instance density, as repeatedly pasting objects often leads to excessive spatial overlap, 
as shown in Table~\ref{tab:aug}. 
Meanwhile, MixUp augmentation is observed to increase the difficulty of localization learning due to blurred object boundaries caused by pixel-level interpolation, as shown in Figure~\ref{fig:giou}. In this paper, we propose GridSynthetic augmentation, which arranges object-centric patches on a multi-scale grid to increase instance density without overlap and preserve clear boundaries for improved localization and vision-text alignment.

\begin{figure*}[t]
    \centering    
    {\includegraphics[width=0.95\textwidth]{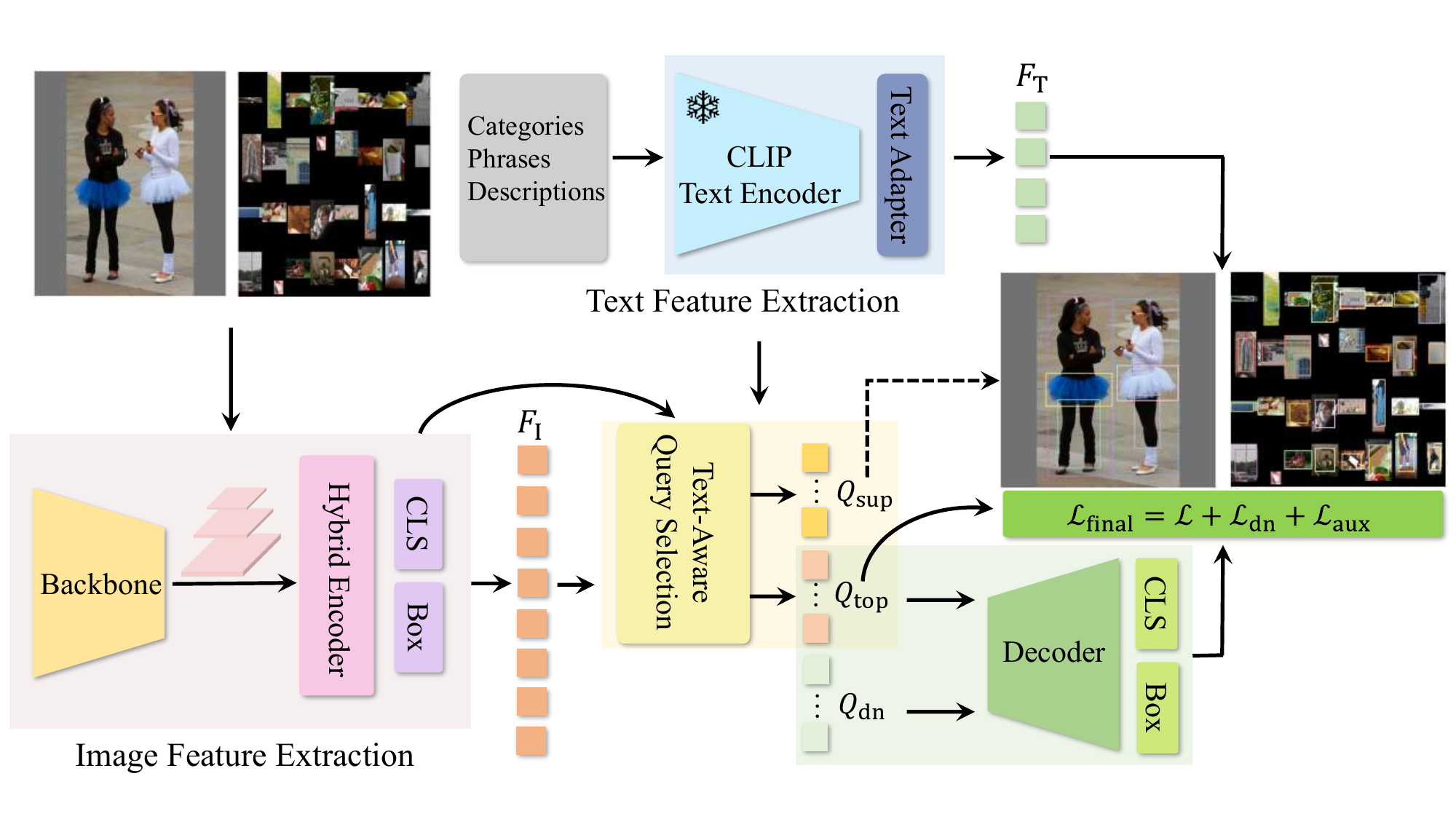}}
    \vspace{-3mm}
    \caption{\textbf{Architecture of the OV-DEIM Framework.} Given an image $\bm{I}$, the backbone and hybrid encoder extract flattened multi-scale visual features $\bm{F_I}$, which are processed by the vision-text alignment head (CLS) to produce similarity scores and into the bounding box regression head to predict object locations. Given text prompts, the text encoder maps them into textual embeddings $\bm{F_T}$, which are used to guide query selection and vision-text alignment. Through the text-aware query selection, the top-ranked queries $Q_{\text{top}}$ are fed into the decoder for iterative refinement, while $Q_{\text{dn}}$ are used for the denoising loss~\cite{denoising} and $Q_{\text{sup}}$ serve as additional queries for computing Fixed AP~\cite{FixedAP}. Each query comprises both a feature embedding and location information.}
    \label{fig:workflow}
\end{figure*}

\section{Background}
In this section, we first revisit DETR-style detectors, focusing on their training and inference procedures, and then present their extension to the open-vocabulary setting.

\subsection{DETR-Style Detectors}
\paragraph{Architecture} 
Given an image $\bm{I} \in \mathbb{R}^{H \times W \times 3}$, where $H$ and $W$ are spatial dimensions and $3$ corresponds to the RGB channels,
a modern DETR-style detector~\cite{rt-detr} employs a backbone network to extract multi-scale features $\{ \bm{S}_3, \bm{S}_4, \bm{S}_5 \}$, where $ \bm{S}_l \in \mathbb{R}^{H_l \times W_l \times D_l}$ with $H_l:=H/2^l$, $W_L:=W/2^l$
denotes the feature map from stage $l$ of the backbone with $D_l$ dimensions. 
These features are then refined and fused through the encoder, obtaining cross-scale features $\bm{F_I} = \{\bm{F}_3, \bm{F}_4, \bm{F}_5\}$,
$ \bm{F}_l \in \mathbb{R}^{H_l \times W_l \times D}$,
and are further flattened into $\bm{F}\in\mathbb{R}^{N\times D}$ with $N=\sum_l H_l W_l$. 
To avoid the heavy complexity of setting each pixel in $\bm{F}$ as an object query, \cite{deformable-detr} employs a query selection mechanism to extract a fixed set of $M$ object queries where $M \ll N$.
The selected queries are iteratively refined by a transformer decoder consisting of self-attention, cross-attention and feedforward network layers.
The head module following the decoder predicts classification scores $\bm{P}\in\mathbb{R}^{M \times C}$, with $C$ being the number of categories, and bounding boxes $\bm{B} \in \mathbb{R}^{N \times 4}$ including the box center coordinates and heights and widths relative to the image size.

\paragraph{Training Process} 
DETR-style detectors formulate object detection as a direct set prediction problem~\cite{detr}. 
During training, object queries are treated as candidate instances and assigned to ground-truth objects through bipartite matching, resulting in a one-to-one correspondence. 
In contrast, YOLO-style detectors~\cite{yoloe,yolo-world} adopt a top-$k$ one-to-many assignment strategy, where each ground-truth instance are matched to multiple candidates, and conflicts are resolved by assigning each candidate exclusively to the ground truth with the highest task-aligned score.
This one-to-many matching scheme can introduce redundant supervision during training.

\paragraph{Inference Process} 
Owing to the one-to-one matching constraint, each object query in DETR-style detectors produces a single prediction consisting of a classification score and a bounding box.
Consequently, 
duplicate predictions for the same object are naturally avoided, and
no category-dependent post-processing is required during the inference.
This leads to lower inference cost which is independent of vocabulary size.
In comparison, the one-to-many matching scheme in YOLO-style detectors leads to highly overlapping predictions, which are resolved using class-wise NMS~\cite{r-cnn,soft-nms}. As a result, the computational cost scales with the number of categories, reducing inference efficiency in large-vocabulary detection.

\subsection{Open-Vocabulary Object Detection}
Given an image $\bm{I}$ and textual prompts $\bm{T}$, which may include category names, noun phrases, or object descriptions, the open-vocabulary object detection (OVOD) task~\cite{glip, yolo-world} aims to localize objects by predicting bounding boxes that match the provided textual descriptions.
During training, beyond learning object localization, the core objective is vision-language alignment. This is achieved by formulating classification as a contrastive learning problem, where the ground-truth text serves as the positive sample and other texts from a candidate pool are treated as negatives~\cite{yolo-world,yoloe}.
OVOD requires aligning each predicted object with a textual description in a large and dynamic vocabulary. This calls for a detection framework that operates at the object level and avoids heuristic components tied to fixed categories. DETR-style detectors meet this requirement by using learnable object queries and one-to-one bipartite matching, which establish a direct correspondence between each query and a single object. This formulation provides a clean interface for vision–language alignment without introducing category-specific design. In addition, by avoiding category-dependent post-processing, they offer more favorable inference scalability as the vocabulary size increases.%

\section{Method}
In this section, we introduce the proposed OV-DEIM framework for open-vocabulary object detection. The OV-DEIM framework is illustrated in Figure~\ref{fig:workflow}. We begin by detailing how vision-language modeling is incorporated into the DETR-style architecture (Section~\ref{sec:arch}), and then present GridSynthetic, a novel data augmentation strategy that strengthens semantic learning for improved vision-language alignment (Section~\ref{sec:gridsyn}).

\subsection{Architecture of the OV-DEIM Framework}
\label{sec:arch}

The overall architecture of the proposed OV-DEIM framework is illustrated in Figure~\ref{fig:workflow}. 
It extends the real-time closed-set detector DEIMv2~\cite{deimv2} to the open-vocabulary setting by incorporating language modeling and vision-text alignment into the detection pipeline. As shown in the figure, OV-DEIM consists of six components: \textit{i)} an image backbone for visual feature extraction, \textit{ii)} a hybrid encoder for multi-scale feature aggregation, \textit{iii)} a text encoder for language representation, \textit{iv)} a text-aware query selection module that identifies high-quality object queries conditioned on language, \textit{v)} a transformer decoder that refines the selected queries for final prediction, and \textit{vi)} a prediction head that outputs vision-text alignment scores and bounding box coordinates.

Among these components, the backbone, encoder, and decoder follow the design of DEIMv2 and are adapted to support language-conditioned detection, specifically
the text-aware query selection module and the vision-text alignment-based classification head, which explicitly integrate textual semantics into query sampling and prediction.

\paragraph{Training Objective}
During training, we optimize a vision-text contrastive loss for classification $\mathcal{L}_{\text{cls}}$, together with an $L_1$ loss $\mathcal{L}_{\text{1}}$ and a GIoU loss~\cite{GIoU} for bounding box regression $\mathcal{L}_{\text{giou}}$. 
All losses are applied to the decoder outputs. 
The overall detection loss is defined as:
\begin{equation*}
\mathcal{L} = \lambda_{\text{cls}} \mathcal{L}_{\text{cls}} + \lambda_{l_1} \mathcal{L}_{\text{1}} + \lambda_{\text{giou}} \mathcal{L}_{\text{giou}},
\end{equation*}
where $\lambda_{\text{cls}}$, $\lambda_{l_1}$, and $\lambda_{\text{giou}}$ balance the contribution of each term.

To stabilize training and improve convergence, we further leverage a denoising loss~\cite{denoising} and an auxiliary loss~\cite{detr}, resulting in the final objective:
\begin{equation*}
\mathcal{L}_{\text{final}} = \mathcal{L} + \mathcal{L}_{\text{dn}} + \mathcal{L}_{\text{aux}}.
\end{equation*}

\paragraph{Inference}
During inference, we introduce a query supplement strategy to improve Fixed AP~\cite{FixedAP} performance by enriching the query set and reducing missed detections. In the following sections, we detail each module and its design.

\paragraph{Text Encoder} 
Given text prompts $\bm{T}$, following YOLOE~\cite{yoloe}, we employ a frozen text encoder from Moble Clip~\cite{mobileclip} to extract pretrained textual embeddings, denoted as $\bm{X}_{T}$.  
Subsequently, a feed-forward block~\cite{glu} is introduced as a text adapter to project the textual embeddings into the visual embedding space, yielding $\bm{F}_T = \text{TextAdapter}(\bm{X}_{T})$, which is similar to YOLOE~\cite{yoloe}.

\begin{figure}[t!]
    \centering
    \includegraphics[width=0.45\textwidth]{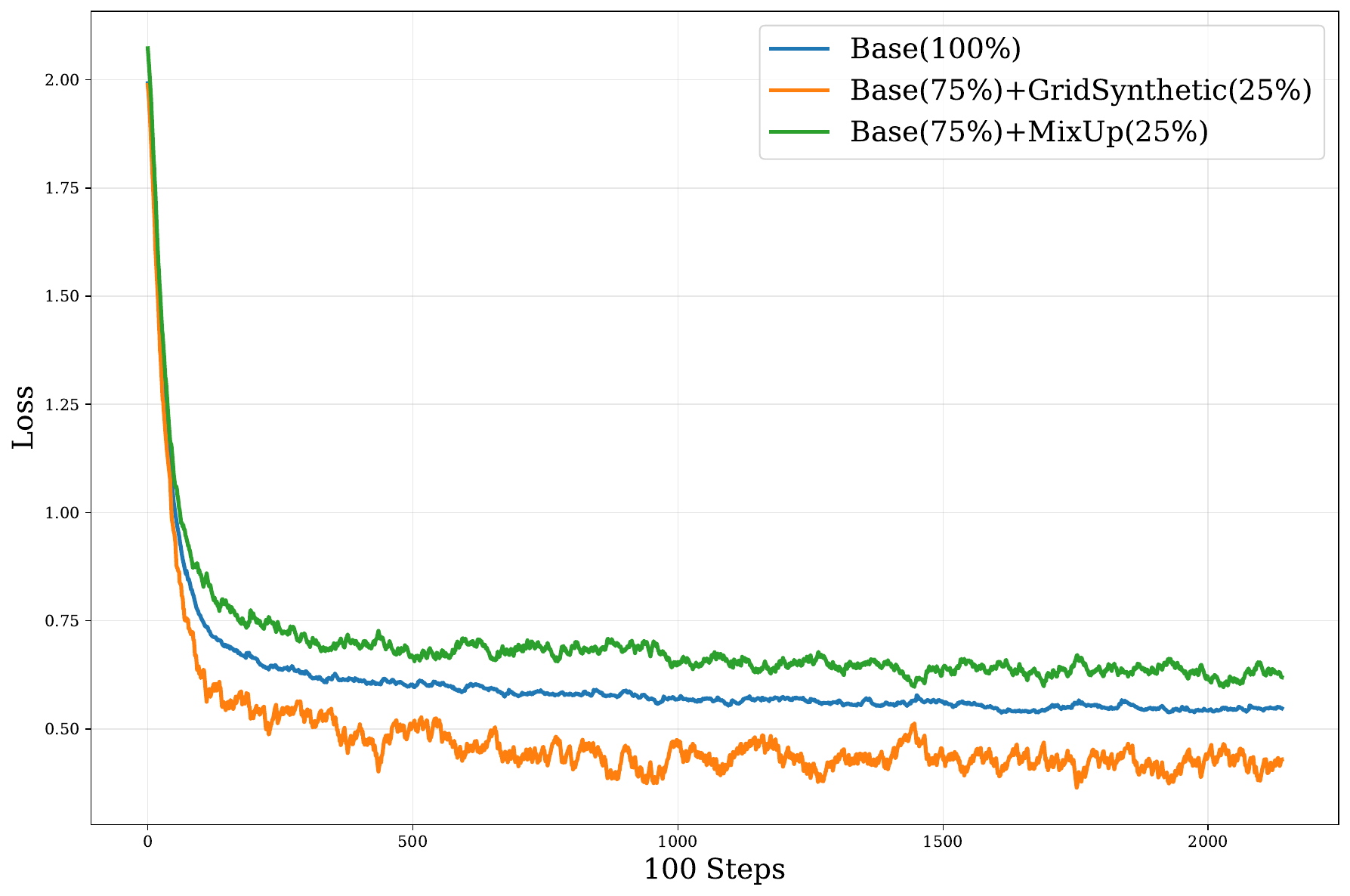}
    \caption{\textbf{Effectiveness of GridSynthetic.} The EMA-smoothed GIoU loss curves show that GridSynthetic consistently achieves the lowest loss throughout training, alleviating the difficulty of localization and leading to more sufficient classification supervision and cleaner semantic alignment.}
    \label{fig:giou}
\end{figure}

\paragraph{Vision-Text Alignment Head} 
The quality of alignment between textual and visual embeddings plays a crucial role in determining the accuracy of category predictions. 
GroundingDINO~\cite{grounding-dino} and OV-DINO~\cite{ov-dino} incorporate cross-modal fusion mechanisms to explicitly integrate visual and textual features, enabling stronger representation alignment between the two modalities.
However, these designs introduce non-negligible computational overhead due to their complex cross-modal interactions, which substantially degrade inference efficiency. 
In contrast, YOLOE~\cite{yoloe} demonstrates that comparable performance can be achieved by discarding such complex designs and instead adopting a lightweight adapter within the text encoder. Therefore, we propose a simple alignment head to compute the vision-text similarity score $p_{i,j}$, which measures the similarity between the $i$-th image feature and the $j$-th class embedding:
\begin{equation*}
p_{i,j}
= \alpha \cdot
\frac{x_i^{\text{img}}}{\lVert x_i^{\text{img}} \rVert_2}
\cdot
\frac{x_j^{\text{text}}}{\lVert x_j^{\text{text}} \rVert_2}
+ \beta ,
\end{equation*}
where the learnable scaling factor $\alpha$ and bias term $\beta$ form an affine transformation for similarity calibration. In particular, $\beta$ is initialized to a large negative value to stabilize training, where negative samples vastly outnumber positive ones.

\paragraph{Text-Aware Query Selection} 
Unlike DETR~\cite{detr}, which directly learns a fixed set of randomly initialized object queries, RT-DETR~\cite{rt-detr} selects the top-$K$ encoder features as initial object queries based on predicted objectness scores. These scores estimate whether a feature corresponds to any foreground object, thereby providing better initialization and reducing the refinement burden of the decoder. However, in the OVOD setting, the objective is to detect those that match the given text prompts. We thus replace objectness-based confidence with vision-text similarity scores for query selection, following OV-DINO~\cite{ov-dino}. Specifically, we rank encoder features according to their similarity with the text embeddings and select the top-$K$ features as object queries. This strategy ensures that the selected queries are not only visually salient but also semantically aligned with the target text prompts, leading to more effective detection in the open-vocabulary setting.

\paragraph{Query Supplement Trick.} Average Precision (AP) evaluates each category independently and does not consider cross-category confidence calibration, which is important in real-world scenarios. Fixed AP~\cite{FixedAP} mitigates this issue by increasing the number of candidate instances per image, thereby improving evaluation performance. However, in DETR-style detectors, the number of candidate predictions is inherently limited by the fixed set of decoder queries. To address this constraint, we introduce a simple query supplement strategy. Specifically, we select additional high-quality queries from the encoder output and treat them as extra detection candidates. This increases the number of predictions per image without modifying the decoder architecture. The strategy is particularly effective for images containing many objects, a common situation in open-vocabulary object detection (OVOD). With this lightweight modification, our method achieves performance comparable to YOLO-style detectors, while keeping the number of decoder queries unchanged and maintaining computational efficiency. %

\paragraph{Vision-Text Contrastive Loss.} In order to assign high classification confidence to well-localized detections, detectors typically incorporate localization-aware signals into the classification loss~\cite{vfl}. Following this principle, we integrate the vision-text similarity score into the MAL loss~\cite{deim} as the classification loss $\mathcal{L}_{\text{cls}}$, replacing the conventional contrastive loss~\cite{SimCLR}. The resulting loss is formulated as follows %
\begin{equation*}
\label{e1}
\mathcal{L}_{\text{cls}} =
\begin{cases}
- \big(q^{\gamma} \log(p) + (1-q^{\gamma}) \log(1-p)\big), & y = 1, \\
- \lambda p^{\gamma} \log(1 - p), & y = 0,
\end{cases}
\end{equation*}
where $p$ denotes the vision-text similarity score predicted by the alignment head, $q$ is the localization quality measured by IoU, $y \in \{0,1\}$ denotes the ground-truth, $\gamma$ is a focusing parameter to emphasize hard negatives, and $\lambda$ controls the contribution of negative samples.

\subsection{Data Augmentation: GridSynthetic}
\label{sec:gridsyn}

\subsubsection{GridSynthetic Design} 

\begin{algorithm}[t!]
\caption{GridSynthetic Data Augmentation}
\label{a1}
\begin{algorithmic}[1]
\INPUT Original dataset $\mathcal{D}$, grid resolution set $\mathcal{G}$, context expansion ratio $\alpha=0.2$, CSS probability $p_{m}=0.5$
\OUTPUT Synthetic samples $\{(I, B, T)\}$ for training

\STATE Construct the object pool $\mathcal{O}$ by extracting object-centric patches from $\mathcal{D}$, preserving category labels and contextual cues (expanded background with ratio $\alpha$)

\FOR{each synthetic sample to generate}
    \STATE Initialize a blank canvas $I$ and sample $(m,n)$ from $\mathcal{G}$
    \STATE Partition $I$ into an $m \times n$ grid
    \STATE Sample $m \times n$ patches from $\mathcal{O}$ and preprocess each patch (crop, flip, resize) to match the grid size
    \STATE Embed each processed patch into a distinct grid cell and record the bounding boxes $B$ and text labels $T$
    \STATE Blend this synthetic sample with another synthetic one using a blend operation with probability $p_{m}$
\ENDFOR
\end{algorithmic}
\end{algorithm}

To facilitate the vision-text alignment learning, the design of GridSynthetic is motivated by the intuition that diverse combinations of object categories should be correctly classified, for capturing robust cross-category semantic relationships, which is summarized in Algorithm~\ref{a1}.

Specifically, we first construct an object pool $\mathcal{O}$ by extracting object-centric patches from the original images, preserving both the associated category and contextual cues, where the context is defined as an expanded background region with an expansion ratio of $0.2$. A blank canvas is then initialized and distinctly partitioned into grids of varying resolutions $(w_m,h_n)$, where $(m,n) \in \{(4,4),(4,8),(8,4),(8,8)\}$, to enable multi-scale object placement. From the object pool $\mathcal{O}$, $m \times n$ objects are sampled and preprocessed through cropping, flipping, and resizing to match the corresponding grid cells. Each processed patch is subsequently embedded into a distinct grid cell to synthesize a composite image containing multiple object instances. Furthermore, to apply complex scene simulation (CSS), each synthetic sample is combined with another synthetic sample using a blend operation with a probability of $0.5$, thereby increasing object density and scene diversity.

\subsubsection{Effectiveness of GridSynthetic}
The operation in GridSynthetic, extracting the object-centric patches and embedding them into a blank canvas, effectively reduces the difficulty of localization and emphasizes semantic recognition during training, as evidenced by the loss curves in Fig.~\ref{fig:giou}. This mechanism mitigates the negative impact of localization inaccuracies on classification learning. According to \eqref{e1}, the positive classification loss in MAL is modulated by the localization quality:
\begin{equation*}
    \mathcal{L}_{\text{pos}} = - \big(q^{\gamma} \log(p) + (1-q^{\gamma}) \log(1-p)\big),
\end{equation*}
 which may cause two issues, insufficient classification supervision and noisy semantic alignment, under the scenario where the localization quality is poor. GridSynthetic, in contrast, produces an idealized scenario in which $q \rightarrow 1$, effectively focusing the learning on the semantic alignment:
\begin{equation*}
    \mathcal{L}_{\text{pos}} = - \log(p).
\end{equation*}

\paragraph{Insufficient Classification Supervision.} When the localization quality is very low, the weighting term approaches zero ($q^{\gamma} \rightarrow 0$), leading to a vanishing gradient with respect to the vision-text similarity score $p$:
\begin{equation*}
    \frac{\partial \mathcal{L}_{\text{pos}}}{\partial p} \propto q^{\gamma} \rightarrow 0.
\end{equation*}
As a result, even when an object of the target category is present in the image, the model is not explicitly guided to align the visual features with the corresponding textual category, limiting effective semantic learning.

\paragraph{Noisy Semantic Alignment.} When the localization quality is moderate rather than precise ($0<q<1$), the loss still explicitly enforces the similarity score $p$ toward 1:
\begin{equation*}
    \arg\min_{p} \mathcal{L}_{\text{pos}}(p, q) \;\; \Rightarrow \;\; p \rightarrow 1.
\end{equation*}
In this case, the model is encouraged to treat region-text pairs from imprecisely localized regions as positive matches. Consequently, textual embeddings are aligned with visual features that include irrelevant background, introducing noise into the learned vision-text representations.

\begin{table}[t!]
\centering
\small 
\setlength{\tabcolsep}{6pt} 
\begin{tabular}{@{}llccc@{}} 
\toprule
Dataset  & Texts & Samples & Images & Annotations \\
\midrule
Objects365V1~\cite{o365}  & 365 & 609k &609k & 9,621k \\
GQA~\cite{gqa}  & 387k & 621k & 46k & 3,681k \\
Flickr30k~\cite{flickr30k} & 94k & 149k & 30k & 641k \\
\bottomrule
\end{tabular}
\vspace{1mm}
\caption{\textbf{Pretraining Datasets.} ``Texts": unique text descriptions; ``Samples": differently annotated images; ``Images": distinct images; ``Annotations": total object-level annotations.}
\label{tab:pretrain_data}
\end{table}
\section{Experiments}

\begin{table*}[t!]
\centering
\small
\setlength{\tabcolsep}{3pt}
\begin{tabular}{lcccccccc}
\toprule
Method & Backbone & Params & Pre-trained Data & FPS & AP/AP$^{\text{Fixed}}$  & AP$_r$/AP$_r$$^{\text{Fixed}}$ & AP$_c$/AP$_c$$^{\text{Fixed}}$ & AP$_f$/ AP$_f$$^{\text{Fixed}}$ \\
\midrule

GLIP-T~\cite{glip} & Swin-T & 232M & OG & - & -/24.9 & -/17.7 & -/19.5 & -/31.0 \\
GLIP-T~\cite{glip} & Swin-T & 232M & OG,Cap4M & - & -/26.0 & -/20.8 & -/21.4 & -/31.0 \\
GLIPv2-T~\cite{glipv2} & Swin-T & 232M & OG,Cap4M & - & -/29.0 & -/- & -/- & -/- \\
GDINO-T~\cite{grounding-dino} & Swin-T & 172M & OG & - & -/25.6 & -/14.4 & -/19.6 & -/32.2 \\
GDINO-T~\cite{grounding-dino} & Swin-T & 172M & OG,Cap4M & - & -/27.4 & -/18.1 & -/23.3 & -/32.7 \\
G1.5-Edge~\cite{GDINO1-5} & EfficientViT-L1 & - & G-20M & - & -/33.5 & -/28.0 & -/34.3 & -/33.9 \\

YOLO-Worldv2-S~\cite{yolo-world} & YOLOv8-S & 13M & OG & - & -/24.4 & -/17.1 & -/22.5 & -/27.3 \\
YOLO-Worldv2-M~\cite{yolo-world} & YOLOv8-M & 29M & OG & - & -/32.4 & -/28.4 & -/29.6 & 35.5 \\
YOLO-Worldv2-L~\cite{yolo-world} & YOLOv8-L & 48M & OG & - & -/35.5 & -/25.6 & -/34.6 & -/38.1 \\

YOLOEv8-S~\cite{yoloe} & YOLOv8-S & 12M & OG & 216/18\textsuperscript{*} & 25.7/27.9 & 19.0/22.3 & 25.9/27.8 & 26.7/29.0 \\
YOLOEv8-M~\cite{yoloe} & YOLOv8-M & 27M & OG & 145/17\textsuperscript{*} & 29.9/32.6 & 23.6/26.9 & 29.2/31.9 & 31.7/34.4 \\
YOLOEv8-L~\cite{yoloe} & YOLOv8-L & 45M & OG & 103/17\textsuperscript{*} & 33.3/35.9 & 30.8/33.2 & 32.2/34.8 & 34.6/37.3 \\

YOLOEv11-S~\cite{yoloe} & YOLOv11-S & 10M & OG & 216/18\textsuperscript{*} & 25.2/27.5 & 19.3/21.4 & 24.4/26.8 & 26.7/29.3 \\
YOLOEv11-M~\cite{yoloe} & YOLOv11-M & 21M & OG & 151/18\textsuperscript{*} & 30.5/33.0 & 22.4/26.9 & 30.4/32.5 & 32.1/34.5 \\
YOLOEv11-L~\cite{yoloe} & YOLOv11-L & 26M & OG & 122/17\textsuperscript{*} & 32.4/35.2 & 25.6/29.1 & 31.9/35.0 & 34.1/36.5 \\

\midrule
OV-DEIM-S & ViT-T & 11M & OG & 161 & 27.7/29.6 & 23.6/25.2 & 28.1/30.2 & 28.0/30.0 \\
OV-DEIM-M & ViT-T+ & 20M & OG & 109 & 30.6/32.6 & 25.3/26.9 & 30.2/31.5 & 31.9/34.1 \\
OV-DEIM-L & ViT-S & 36M & OG & 91 & 33.7/35.9 & 34.3/36.8 & 33.4/35.5 & 34.0/36.0 \\
\bottomrule
\end{tabular}
\vspace{1mm}
\caption{\textbf{Zero-shot Evaluation on LVIS~\cite{lvis}.} AP denotes the standard Average Precision, whereas AP$^{\text{Fixed}}$ corresponds to Fixed AP, both evaluated on the LVIS \texttt{minival} split. Inference speed (FPS) is measured on an NVIDIA T4 GPU using TensorRT. Following RT-DETR~\cite{rt-detr}, FPS marked with * includes the post-processing overhead of NMS, averaged over the 5,000 images in the \texttt{minival} set. OG indicates Objects365v1~\cite{o365} and GoldG~\cite{goldg}, Cap4M is a collection of 4M image-text pairs from GLIP~\cite{glip}, and G-20M represents Grounding-20M~\cite{GDINO1-5}.}
\label{tab:zeroshot_lvis}
\end{table*}

\subsection{Experimental Setup}

\paragraph{Implementation Details.}

Following~\cite{yolo-world}, we provide three OV-DEIM variants to accommodate different latency requirements: small (S), medium (M), and large (L). Following DEIMv2~\cite{deimv2}, the backbone is based on DINOv3~\cite{dinov3} with the Spatial Tuning Adapter (STA), while the encoder and decoder are adopted from RT-DETR~\cite{rt-detr}. For fair comparisons with YOLOE~\cite{yoloe}, text prompts are processed using the pretrained MobileCLIP-B(LT)~\cite{mobileclip} text encoder. The S and M variants are trained for 30 epochs on 8 NVIDIA RTX 4090 GPUs with a total batch size of 128, leveraging cached text embeddings and removing the text encoder during training, whereas the L variant and all ablation experiments on OV-DEIM-L are trained for 20 epochs. Notably, the number of decoder queries is fixed at 300 to meet real-time requirements. We further introduce 700 additional queries as part of the query supplement strategy, resulting in a total of 1000 bounding box predictions per image. Moreover, $\alpha$ and $\beta$ are set to $\ln (15)$ and $-\ln (100)$, respectively. We set $\lambda = 0.5$ in the classification loss $\mathcal{L}_{\text{cls}}$ to place greater emphasis on positive samples. All other hyperparameters follow DEIMv2~\cite{deimv2}.

During training, we apply color augmentation, random affine transformations, random flipping, and mosaic with a probability of 0.75, while GridSynthetic and MixUp augmentations are applied with probabilities of 0.125 each.

\paragraph{Training Data.} To enable open-vocabulary detection, we use the Objects365v1~\cite{o365} detection dataset along with grounding datasets including GoldG~\cite{goldg} (GQA~\cite{gqa} and Flickr30k~\cite{flickr30k}), as summarized in Table~\ref{tab:pretrain_data}. Consistent with~\cite{glip}, the COCO dataset is excluded from GoldG~\cite{goldg} (GQA and Flickr30k). Note that this follows the standard evaluation protocol used in YOLO-World~\cite{yolo-world} and YOLOE~\cite{yoloe}.

\paragraph{Zero-shot Evaluations.} After pretraining, we evaluate the zero-shot performance of OV-DEIM on the LVIS~\cite{lvis} dataset and the COCO~\cite{coco} dataset. The LVIS dataset contains 1,203 object categories, which is suitable for evaluating performance on large-vocabulary and long-tailed scenarios. For comparison, we evaluate on the LVIS \texttt{minival} split~\cite{yolo-world}. We report the standard AP with the maximum number of predictions per image set to 300, and Fixed AP~\cite{FixedAP} with the maximum number of predictions per image increased to 1,000. In contrast, COCO contains 80 object categories that are dominated by common classes with relatively balanced distributions. We evaluate on the \texttt{val2017} split using the standard COCO AP metrics.

\subsection{Comparison with State-of-the-art Approaches}

\paragraph{Zero-shot Results on LVIS.}

As shown in Table~\ref{tab:zeroshot_lvis}, OV-DEIM surpasses previous state-of-the-art methods in both zero-shot performance and inference speed. Despite using fewer model parameters, OV-DEIM pretrained on OG outperforms GLIP variants and Grounding DINO models, which leverage substantially larger pretraining datasets. Compared to YOLO-style detectors, OV-DEIM-S/M/L outperform YOLOEv8-S/M/L by 2.0/0.7/0.4 AP, while achieving comparable Fixed AP, together with 8.9$\times$/6.4$\times$/5.4$\times$ inference speedups on an NVIDIA T4 GPU.
Moreover, on the challenging rare-category split, OV-DEIM-S and OV-DEIM-L achieve notable improvements of 31.09\% and 11.36\% in AP$_r$, together with gains of 14.80\% and 10.84\% in Fixed AP$_r$, respectively. OV-DEIM-M attains comparable performance on Fixed AP$_r$ while improving AP$_r$ by 7.21\%. %
These results demonstrate the strong generalization of OV-DEIM under long-tailed distributions. %

\paragraph{Zero-shot Results on COCO.}

\begin{table}[t!]
    \centering
    \small
    \setlength{\tabcolsep}{1pt}
    \begin{tabular}{lccccc}
    \toprule
        Method & Backbone & Params & AP & AP$_{50}$ & AP$_{75}$ \\
    \midrule
    \textit{Zero-shot transfer} \\
    
        YOLO-Worldv1-S~\cite{yolo-world} & YOLOv8-S & 13M & 37.6 & 52.3 & 40.7  \\
        YOLO-Worldv1-M~\cite{yolo-world} & YOLOv8-M & 29M & 42.8 & 58.3 & 46.4  \\
        YOLO-Worldv1-L~\cite{yolo-world} & YOLOv8-L & 48M & 44.4 & 59.8 & 48.3  \\
    \midrule
    \textit{Linear probing} \\
        YOLOEv8-S~\cite{yoloe} & YOLOv8-S & 12M & 35.6 & 51.5 & 38.9  \\
        YOLOEv8-M~\cite{yoloe} & YOLOv8-M & 27M & 42.2 & 59.2 & 46.3 \\
        YOLOEv8-L~\cite{yoloe} & YOLOv8-L & 45M & 45.4 & 63.3 & 50.0 \\
        YOLOEv11-S~\cite{yoloe} & YOLOv11-S & 10M & 37.0 & 52.9 & 40.4 \\
        YOLOEv11-M~\cite{yoloe} & YOLOv11-M & 21M & 43.1 & 60.6 & 47.4 \\
        YOLOEv11-L~\cite{yoloe} & YOLOv11-L & 26M & 45.1 & 62.8 & 49.5 \\
    \midrule
    \textit{Zero-shot transfer} \\
        OV-DEIM-S & ViT-T & 11M & 40.8 & 56.3 & 44.4 \\
        OV-DEIM-M & ViT-T+ & 20M & 43.3 & 60.2 & 48.0 \\
        OV-DEIM-L & ViT-S & 35M & 45.9 & 62.3 & 49.9 \\
    \bottomrule
    \end{tabular}
    \vspace{1mm}
    \caption{\textbf{Zero-shot Evaluation on COCO~\cite{coco}.} The proposed model achieves better zero-shot performance on COCO compared to YOLO-Worldv1-S~\cite{yolo-world}. It also consistently outperforms the linear probing version of YOLOE~\cite{yoloe} in zero-shot evaluation.}
    
    \label{tab:coco}
\end{table}

\begin{figure*}[t]
    \centering    
    {\includegraphics[width=0.95\textwidth]{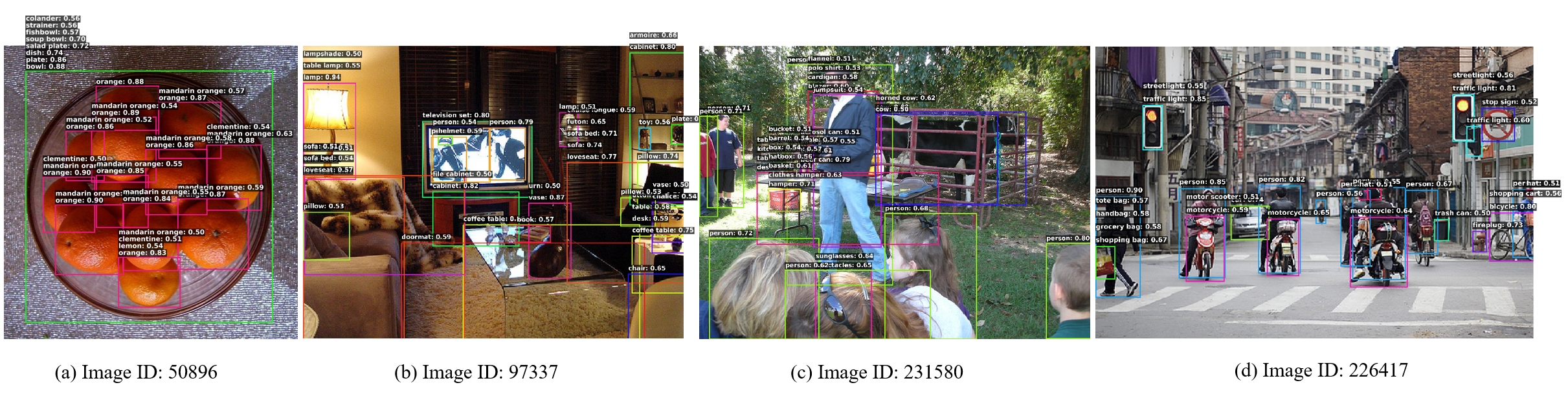}}
    \vspace{-3mm}
    \caption{\textbf{Visualizations of Zero-shot Inference on LVIS~\cite{lvis}.} We employ the pretrained OV-DEIM-L model and perform inference with the LVIS vocabulary containing 1,203 categories, only using a confidence threshold of 0.5.}
    \label{fig:four_images}
    
\end{figure*}

\begin{table}[t!]
\centering
\setlength{\tabcolsep}{4pt}
\begin{tabular}{lcccc}
\toprule
Data Augmentation  & AP & AP$_r$ & AP$_c$ & AP$_f$  \\

\midrule
\multicolumn{5}{c}{\textbf{Ablation on GridSynthetic}} \\
\midrule
Base w/ GS (full) & 33.1 & 29.3 & 33.5 & 33.4 \\
\qquad (w/o CSS) & 32.5 & 30.2 & 32.2 & 33.1 \\
\qquad (w/o CSS \& Context Cues) & 30.6 & 29.3 & 31.3 & 30.3 \\

\midrule
\multicolumn{5}{c}{\textbf{General Data Augmentations}} \\
\midrule
Base & 32.4 & 29.3 & 32.6 & 32.7 \\
Base w/ CopyPaste$_{4}$~\cite{copy_paste} & 32.1 & 29.3 & 32.6 & 32.2 \\
Base w/ CopyPaste$_{16}$~\cite{copy_paste} & 32.5 & 25.3 & 34.1 & 32.2 \\
Base w/ MixUp~\cite{mixup_od} & 33.1 & 29.4 & 33.2 & 33.6 \\
Base w/ GS & 33.1 & 29.3 & 33.5 & 33.4 \\
Base w/ GS \& MixUp~\cite{mixup_od} & 33.7 & 34.3 & 33.4 & 34.0 \\
\bottomrule
\end{tabular}
\vspace{1mm}
\caption{
\textbf{Ablation Study on Data Augmentations and GridSynthetic.} GS denotes GridSynthetic, CSS represents complex scene simulation, and CopyPaste$_{4/16}$ indicates pasting 4 or 16 objects per sample. Base augmentations (color, affine, flip, mosaic) are applied with probability 0.75, while additional augmentations (MixUp and/or GridSynthetic) are applied with probabilities adjusted per variant, \textit{e.g.}, 0.25 for Base w/ MixUp, 0.125 each for Base w/ GS \& MixUp. All experiments are conducted using OV-DEIM-L and trained on the OG dataset.}
\label{tab:aug}
\end{table}

As illustrated in Table~\ref{tab:coco}, OV-DEIM demonstrates strong zero-shot transfer capability on COCO, which comprises 80 object categories, most of which are common and relatively evenly distributed. Across all model scales, OV-DEIM consistently outperforms YOLO-World in zero-shot transfer, with OV-DEIM-S/M/L surpassing them by 3.4/0.5/1.5 AP, respectively. Similarly, when compared to the linear-probing results of YOLOE, OV-DEIM-S/M/L achieve gains of 4.0/0.2/0.7 AP, demonstrating that its performance advantage is not limited to zero-shot transfer but also holds against task-adapted baselines. Moreover, OV-DEIM demonstrates consistent improvements in AP$_{75}$ across all scales, reflecting more accurate high-IoU localization, which can be attributed to the iterative refinement performed by the decoder module.

\subsection{Ablation Study}

In this section, we validate our design choices. Unless otherwise specified, all experiments are conducted on the OG dataset, which combines Objects365v1~\cite{o365} and GoldG~\cite{goldg}. We report results on LVIS~\cite{lvis} using the OV-DEIM-L.

\paragraph{Impact of GridSynthetic.} Table~\ref{tab:aug} validates the effectiveness of GridSynthetic. Each component contributes positively to detection performance. Removing CSS reduces AP from 33.1 to 32.5, indicating that simulating more complex scenes improves robustness. Further removing contextual expansion causes a larger drop to 30.6 AP across all splits, showing that preserving boundary-aware context around object-centric patches provides richer spatial cues for more accurate localization, rather than artificially simplifying the task. In comparison, naive instance-level augmentation such as Copy-Paste brings limited gains and can even hurt performance—while CopyPaste$_{16}$ slightly improves AP on common categories, it significantly degrades rare-category results, likely due to excessive spatial overlap and biased supervision. MixUp consistently improves overall AP. Notably, combining GridSynthetic with MixUp yields the best overall performance and substantially improves rare-category detection, demonstrating that the two strategies are complementary and reinforcing.

\paragraph{Impact of Query Supplement Trick.} Table~\ref{tab:extra_queries} demonstrates the effectiveness of the proposed query supplement strategy. In OVOD, images often contain many objects, and the Fixed AP metric benefits from a larger set of detection candidates. By incorporating additional encoder queries, we increase the number of predicted boxes without modifying the decoder. As the number of extra queries increases, AP consistently improves from 33.1 to 35.3, confirming that expanding the candidate set enhances evaluation performance. The gains are consistent across AP$_r$, AP$_c$, and AP$_f$, indicating improvements for rare, common, and frequent categories alike. We also observe that performance gains gradually saturate beyond 400 extra queries, suggesting that further expansion brings limited additional benefit. Importantly, the query supplement strategy introduces no extra inference latency, since encoder predictions remain NMS-free and do not increase decoder computation.

\begin{table}[t!]
    \centering
    \setlength{\tabcolsep}{5pt}
    \begin{tabular}{lcccc}
    \toprule
       Extra Queries   & AP$^{\text{Fixed}}$ & AP$_r$$^{\text{Fixed}}$ & AP$_c$$^{\text{Fixed}}$ & AP$_f$$^{\text{Fixed}}$ \\
    \midrule
        0 & 33.1 & 30.3 & 33.0 & 33.7 \\
        100 & 33.7 & 30.8 & 33.6 & 34.4 \\
        200 & 34.2 & 30.8 & 34.1 & 34.9 \\
        300 & 34.5 & 31.0 & 34.6 & 35.2 \\
        400 & 34.8 & 31.4 & 34.8 & 35.4 \\
        500 & 34.9 & 31.5 & 35.0 & 35.6 \\
        600 & 35.2 & 32.1 & 35.2 & 35.7 \\
        \bf 700 & \bf 35.3 & \bf 32.2 & \bf 35.3 & \bf 35.8 \\
    \bottomrule
    \end{tabular}
    \vspace{1mm}
    \caption{
    \textbf{Ablation Study on Query Supplement Trick.} Beyond the default 300 decoder queries, we introduce additional candidate queries from the encoder to generate more bounding box predictions. This simple strategy increases the number of detection candidates per image and proves effective under the Fixed AP~\cite{FixedAP}.
    }
    \label{tab:extra_queries}
\end{table}

\paragraph{Visualization.} Figure~\ref{fig:four_images} illustrates qualitative detection results on LVIS categories generated by OV-DEIM-L with a confidence threshold of 0.5, highlighting two key advantages. First, our OV-DEIM-L accurately localizes objects even in crowded scenes and for small-scale instances, demonstrating strong robustness in challenging spatial configurations. Second, each predicted bounding box is associated with multiple semantically relevant text prompts, indicating effective vision-text alignment and the ability to capture rich, fine-grained semantic concepts.

\section{Conclusion}
In this work, we present OV-DEIM, a real-time DETR-style detector for open-vocabulary object detection. By extending DEIM to the open-vocabulary setting, OV-DEIM leverages one-to-one matching to eliminate NMS post-processing, enabling fast and stable inference. To overcome the fixed-query limitation of DETR-style models, we introduce a query supplement trick that enlarges the candidate pool using additional encoder queries, improving Fixed AP without increasing inference cost. To strengthen semantic recognition, we further propose GridSynthetic, a data augmentation strategy that enhances supervision and is complementary to MixUp. By constructing idealized localization scenarios, GridSynthetic reduces localization-induced noise in the classification loss and promotes more robust cross-category semantic alignment. Extensive experiments show that OV-DEIM achieves strong zero-shot performance on LVIS and COCO, especially on rare categories. We hope that it can serve as a strong baseline for future research in real-time open-vocabulary object detection.

\bibliography{example_paper}
\bibliographystyle{ieeetr}

\end{document}